\documentclass{article}

\usepackage{microtype}
\usepackage{graphicx}
\usepackage{subfigure}
\usepackage{booktabs} % for professional tables

\usepackage{tabularx}
\usepackage{varwidth}
\usepackage{afterpage}

\usepackage{hyperref}

\usepackage[accepted]{icml2025}

\usepackage{amsmath}
\usepackage{amssymb}
\usepackage{mathtools}
\usepackage{amsthm}

\usepackage[capitalize,noabbrev]{cleveref}

\theoremstyle{plain}

\theoremstyle{definition}

\theoremstyle{remark}

\newcommand{\WW}{\boldsymbol{W}}
\newcommand{\bb}{\boldsymbol{b}}
\newcommand{\xx}{\boldsymbol{x}}

\usepackage[textsize=tiny]{todonotes}

\icmltitlerunning{Transcoders Beat Sparse Autoencoders for Interpretability}

\begin{document}

\twocolumn[
\icmltitle{Transcoders Beat Sparse Autoencoders for Interpretability}



\icmlsetsymbol{equal}{*}

\begin{icmlauthorlist}
\icmlauthor{Gonçalo Paulo}{equal,eai}
\icmlauthor{Stepan Shabalin}{equal,eai}
\icmlauthor{Nora Belrose}{equal,eai}
\end{icmlauthorlist}

\icmlaffiliation{eai}{EleutherAI}

\icmlcorrespondingauthor{Gonçalo Paulo}{goncalo@eleuther.ai}
%\icmlcorrespondingauthor{Firstname2 Lastname2}{first2.last2@www.uk}

\icmlkeywords{Machine Learning, ICML}

\vskip 0.3in
]



\printAffiliationsAndNotice{\icmlEqualContribution} % otherwise use the standard text.

\begin{abstract}
Sparse autoencoders (SAEs) extract human-interpretable features from deep neural networks by transforming their activations into a sparse, higher dimensional latent space, and then reconstructing the activations from these latents. Transcoders are similar to SAEs, but they are trained to reconstruct the output of a component of a deep network given its input. In this work, we compare the features found by transcoders and SAEs trained on the same model and data, finding that transcoder features are significantly more interpretable. We also propose \emph{skip transcoders}, which add an affine skip connection to the transcoder architecture, and show that these achieve lower reconstruction loss with no effect on interpretability. %on also use a  the Pareto curve of interpretability and performance for SAEs, transcoders, and skip transcoders, as well as recently developed benchmarks measuring the capabilities related to probing, unlearning and absorption.
\end{abstract}

\section{Introduction}

Recently, large language models have achieved human-level reasoning performance in many tasks \citep{guo2025deepseek}. Interpretability aims to improve the safety and reliability of these systems by understanding their internal mechanisms and representations. While early research attempted to produce natural language explanations of individual neurons \citep{Olah2020,gurnee2023finding,gurnee2024universal}, it is now widely recognized that most neurons are ``polysemantic'', activating in semantically diverse contexts \citep{arora2018linear,elhage2022toy}.

Sparse autoencoders (SAEs) have emerged as a promising tool for partially overcoming polysemanticity, by decomposing activations into interpretable features \cite{bricken2023monosemanticity,templeton2024scaling,gao2024scaling}. SAEs are single hidden layer neural networks trained with the objective of reconstructing activations with a sparsity penalty \cite{bricken2023monosemanticity,rajamanoharan2024jumping}, sparsity constraint \cite{gao2024scaling, bussmann2024batchtopksparseautoencoders}, or an information bottleneck \cite{ayonrinde2024interpretability}. They consist of two parts: an encoder that projects activations into a sparse, high-dimensional latent space, and a decoder that reconstructs the original activations from the latents.

\citet{bricken2023monosemanticity} introduced a technique of evaluating the interpretability of SAEs by simulating them with an LLM-based scorer, similar to what had been done on neurons \cite{bills2023language}.
This approach is commonly called automated interpretability, or autointerp. SAE features perform much better on this benchmark compared to neurons, even when neurons are ``sparsified'' by selecting only the top-$k$ most active neurons in a layer for analysis \citep{paulo2024automatically}. One problem with SAEs is that they focus on compressing intermediate activations rather than modeling the \textit{functional behavior} of network components (e.g., feedforward modules).

\begin{figure*}[!t]
    \centering    \includegraphics[width=1\linewidth]{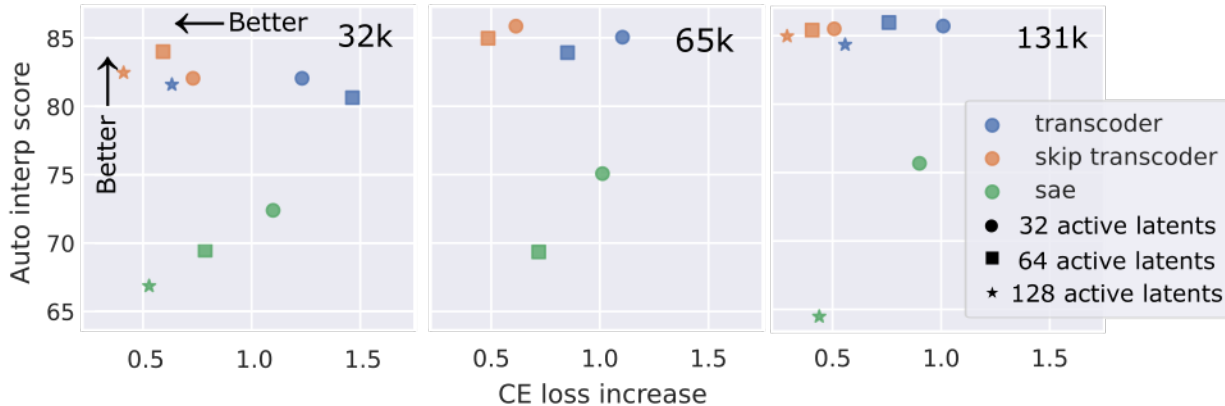}
    \vskip -0.3cm
    \caption{\textbf{Skip transcoders are a Pareto improvement on interpretability vs performance degradation}.
    We compare the increase in cross-entropy loss of 3 different sizes of SAEs and transcoders, 32768  (top right), 65536  (bottom left) and 131072 (bottom right), when patched into the model. For all sizes, skip transcoders are better than transcoders and sparse autoencoders, having both lower increase in model loss and a higher average auto interpretability score. On each quadrant we show 3 models that were trained with a different number of active latents, 32, 64 and 128, except for the 65536 latent model, which only has 32 and 64. The auto interp score is defined as the average fuzzing and detection score of c.a. 500 latents.   
    }
    \label{fig:pareto}
\end{figure*}

\textbf{Transcoders} are an alternative to sparse autoencoders, initially proposed in \citet{li2023dictionarylearning} and \citet{templeton2024predicting}, and first rigorously evaluated by \citet{transcoders_paper}. Unlike SAEs, transcoders approximate the \textit{input-output function} of a target component, such as an an MLP, using a sparse bottleneck. \citet{transcoders_paper} demonstrate that transcoders enable fine-grained circuit analysis by learning input-invariant descriptions of component behavior, complementing automated circuit discovery tools like \citet{conmy2023automatedcircuitdiscoverymechanistic}.

% One way to evaluate SAEs is to measure the increase in a model's next-token prediction loss when an SAE is ``patched in,'' replacing the true intermediate activations with their reconstructed counterparts. The analogous evaluation method for a transcoder is to replace the

\begin{figure*}[t]
    \centering    \includegraphics[width=1\linewidth]{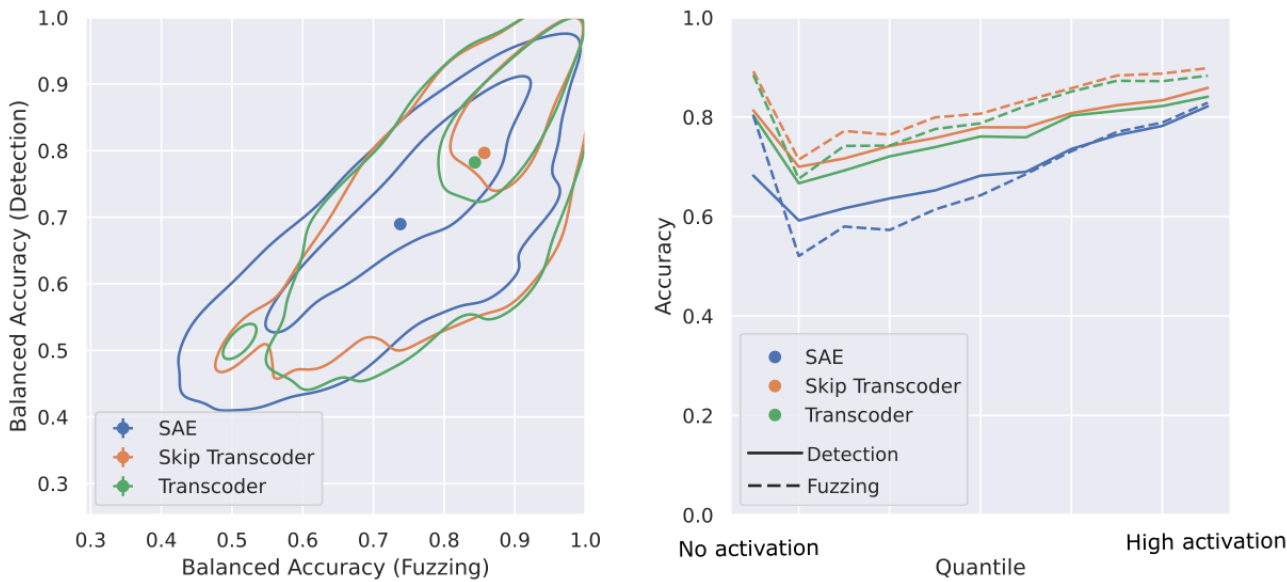}   
    \vskip -0.3cm \caption{\textbf{Interpretability of latents and generalization of explanations.} The interpretability scores of both detection and fuzzing are higher for skip transcoders and transcoders when compared to SAEs, with the distribution being wider for SAEs. Dots in the left plot indicate the average score. The accuracy of the explanations on examples sampled from different quantiles of the activation distribution we can observe that 
    The accuracy of explanations remains higher even for lower quantiles, where the activations are smaller, showing that transcoder and skip-transcoder latents are probably representing more monosemantic concepts along the full distribution. 
    }
    \label{fig:distribution}
\end{figure*}

Transcoder design faces inherent challenges. While ReLU MLPs carve up the input space into polytopes,\footnote{The polytope interpretation can also be applied, in a somewhat modified form, to MLPs with other activation functions \citep{balestriero2018from}.} with each polytope corresponding to a relatively high-rank linear function \citep{black2022interpretingneuralnetworkspolytope}, transcoders' sparse activations mean that each activation pattern corresponds to a low-rank linear transformation. Furthermore, since \citet{marks2024sparse} and \citet{bricken2023towards}, new benchmarks for sparse feature evaluation have emerged \cite{gao2024scaling,karvonen2024saebench,juang2024autointerp}, motivating a broader evaluation of transcoders across models and tasks. We investigate the tradeoff between reconstruction error and interpretability by comparing SAEs and transcoders and address challenges mentioned above by proposing an architectural improvement—the \textbf{skip transcoder}—which mitigates rank limitations via an affine skip connection.

In this work, we:
\begin{enumerate}
    \item Introduce skip transcoders, which reduce reconstruction error without compromising interpretability.
    \item Compare transcoders, skip transcoders, and SAEs across diverse models (up to 2B parameters), showing skip transcoders Pareto-dominate SAEs on reconstruction vs. interpretability tradeoffs.
    \item Evaluate transcoders on SAEBench \cite{karvonen2024saebench} demonstrating improved quality in both latent-level phenomena like absorption and performance on various tasks through sparse probing.
\end{enumerate}

We conclude that interpretability researchers should shift their focus away from sparse autoencoders trained on the outputs of MLPs and toward (skip) transcoders.

\section{Methods}

Skip transcoders add a linear ``skip connection'' to the transcoder, which we find improves its ability to approximate the original MLP at no cost to interpretability scores. Specifically, the transcoder takes the functional form
\begin{equation}
    f(\xx) = \WW_2 \mathrm{TopK}(\WW_1 \xx + \bb_1) + \WW_{\mathrm{skip}} \xx + \bb_2
\end{equation}
Both $\WW_2$ and $\WW_{\mathrm{skip}}$ are zero-initialized, and $\bb_2$ is initialized to the empirical mean of the MLP outputs, so that the transcoder is a constant function at the beginning of training. We leave a deeper analysis of the skip connection, perhaps interpreting it using SVD \citep{millidge2022singular}, for future work.

\subsection{Training}

We train a collection of sparse coders: sparse autoencoders (SAE), sparse transcoders (ST), and sparse skip transcoders (SST), on the MLP layers of Pythia 160M \cite{biderman2023pythia}. We also train SAEs and SSTs on Llama 3.2 1B and Gemma 2 2B. We train with mean squared error between the output of the sparse coder and the MLP output, with no auxiliary loss terms. Unlike prior work on transcoders, we adopt the state-of-the-art TopK activation function proposed by \citet{gao2024scaling}, which directly enforces a desired sparsity level on the latent activations without the need to tune an L1 sparsity penalty. We sweep across $k$ values of 32, 64, and 128 in our experiments.

For sparse coders trained on Pythia, we train over the first 8B tokens of Pythia's training corpus, the Pile \citep{gao2020pile}. For the other models, we use 8B tokens of the RedPajama v2 corpus \citep{together2023redpajama}. All sparse coders are trained using the Adam optimizer \citep{kingma2014adam}, a sequence length of 2049, and a batch size of 64 sequences.

\subsection{Evaluation}

We use the automated interpretability pipeline released by \citet{paulo2024automatically} to generate explanations and scores for sparse coder latents. Activations of latents were collected over 10M tokens, sampled from the Pile for the Pythia models and from FineWeb \citep{penedo2024the} for Llama and Gemma. The explanations were generated by showing an explainer model, Llama 3.1 70b, 40 activating examples, four from each of ten different quantiles. Each example had 32 tokens, and the active tokens were highlighted. Detection and fuzzing scores were computed over 50 activating examples, five from each of ten different quantiles, and 50 non-activating examples. Simulation scores were computed over the same 50 activating examples, as described in \cite{bills2023language}. The scorer model was Llama 3.1 70b.

We also use the SAEBench repository \cite{karvonen2024saebench} to evaluate the sparse coders. We use it to compute the variance explained and the cross-entropy loss increase over 500K tokens of the OpenWebText corpus \citep{Gokaslan2019OpenWeb}. SAEBench also provides the ability to train and evaluate \textbf{sparse probes} that measure the ability of the SAE's encoder to select information relevant to classification tasks such as sentiment and language detection.

Recently, \citet{chanin2024absorption} drew attention to the phenomenon of \textbf{feature absorption}. In some cases, a more general feature like \emph{starts with the letter L} appears alongside a specific feature like \emph{the token ``lion''}, which may prevent the general feature from being active in contexts where intuitively, both the general and the specific feature apply. They argue that this is undesirable. We use SAEBench to compute the frequency of absorption of general letter features into specific features, in SAEs, STs, and STSs.

\begin{table*}[htbp]
\centering
\begin{tabular}{|c|c|c|ccc|ccc|ccc|ccc|}
\hline
 & &  & \multicolumn{3}{c|}{\textbf{Fuzzing (\%, $\uparrow$)}} & \multicolumn{3}{c|}{\textbf{Detection (\%, $\uparrow$)}} & \multicolumn{3}{c|}{\textbf{Simulation ( $\uparrow$)}} & \multicolumn{3}{c|}{\textbf{CE Loss Increase (\%, $\downarrow$)}} \\
\textbf{Model} & \textbf{Size} & K & SAE & ST & SST & SAE & ST & SST & SAE & ST & SST & SAE & ST & SST \\
\hline
pythia-160m & $2^{15}$ & 32 & 74.6 & 85.4 & \textbf{86.4} & 70.2 & 78.7 & \textbf{80.9} & 0.28 & 0.46 & \textbf{0.47} & 1.10 & 1.23 & \textbf{0.73}\\
pythia-160m &  $2^{15}$  & 64 & 71.8 & 83.7 & \textbf{86.9} & 67.2 & 77.5 & \textbf{81.1} & 0.30 & - & \textbf{0.42} & 0.79 & 1.46 & \textbf{0.59}\\
\hline
pythia-410m &  $2^{16}$ & 32 & 78.4 & - & \textbf{89.5} & 72.7 & - & \textbf{83.8} & 0.35 & - & \textbf{0.51} & 0.49 & - & \textbf{0.43}\\
\hline
llama-1b &  $2^{17}$ & 32 & 77.2 & - & \textbf{85.7} & 71.7 & - & \textbf{79.4} & 0.34 & - & \textbf{0.44} & 1.50 & - & \textbf{1.00}\\
\hline
gemma-2-2b & $2^{17}$  & 32 & 80.5 & - & \textbf{84.6} & 75.8 & - & \textbf{79.6} & 0.35 & - & \textbf{0.44} & 1.60 & - & \textbf{0.53}\\
\hline
\end{tabular}
\caption{\textbf{Performance of sparse coders on different models} We compute different interpretability scores, fuzzing, detection and simulation, for SAEs and SSTs trained on different models, as well as the increase of cross-entropy loss when patched into the model. 500 latents are used for fuzzing and detection, but only 50 latents are used for simulation due to it being more computationally expensive. 0.5M tokens are used to compute the cross-entropy loss increase.}
\label{tab:models}
\end{table*}

\section{Skip Transcoders Pareto Dominate SAEs}

The utility of any sparse coding method for interpretability lies in its ability to accurately reconstruct activations while also generating human-interpretable latent features. This is a fundamental tradeoff: while sparser latents are generally more interpretable, higher sparsity also tends to increase the reconstruction error. The reconstruction error of a sparse coder can be viewed as ``dark matter'' containing features not captured by the latents \citep{engels2024decomposing}.

Following earlier work on SAEs, we can represent this tradeoff using a reconstruction vs. interpretability curve (Figure~\ref{fig:pareto}). Here we compare the reconstruction loss of different models, varying the number number of latents and sparsity, with their interpretability scores, measured as the average between detection and fuzzing score over a set of features. We find transcoders and skip transcoders with the same number of latents generally have higher interpretability scores for the same reconstruction loss than SAEs.

Not only are the average interpretability scores of transcoders and skip transcoders higher than those of SAEs but their distribution is narrower (Figure~\ref{fig:distribution}, left panel). Latents of (skip) transcoders also seem to represent more monosemantic features, as the explanations found hold for larger portion of the activation distribution. This can be seen by comparing the accuracy of explanations in examples sampled from different quantiles of the activation distribution (Figure~\ref{fig:distribution}, left). The accuracy of explanations decreases more slowly for STs and SSTs than SAEs. The explanations are also more sensitive, as the false positive rate is lower.

We replicated these results in models of the same architecture but different sizes, Pythia~160m and Pythia~410m, and on larger models with different architectures, Llama 3.2 1B \citep{dubey2024llama} and Gemma 2 2B \citep{team2024gemma}. On all cases studied, SSTs had higher automated interpretability scores and lower CE loss increase when patched in, see Table \ref{tab:models}. 

We found that performance on sparse probing was similar for SSTs and SAEs (\cref{sec:app-saebench}), with SSTs winning out for later layers by a small margin. Sparse probing measures the ability of SAEs to preserve information in the original latent, but for transcoders the latents should relate more to concepts necessary for processing the input. It is thus surprising that they are competitive with SAEs on compressing the residual stream without being trained with that objective. We also find that SAEs and transcoders have similar  feature absorption behavior, but that those results are noisy. We don't expect this to be a problem, since there are other methods orthogonal to ours which seem to improve feature absorption; see discussion in \cref{sec:future}.

\section{Conclusion}

Our experiments suggest that interpretability researchers should shift their focus from sparse autoencoders trained in the outputs of MLPs to (skip) transcoders.
In our view, the only downside of transcoders compared to sparse autoencoders is that SAEs can be trained directly on the residual stream, while transcoders need to be trained on particular components of the model (usually a feedforward layer). However, one can easily convert a skip transcoder trained on an FFN into a ``residual stream transcoder'' by adding the identity matrix to its skip connection. In this way, skip transcoders can be viewed as bridging the gap between these two types of sparse coding. Additionally, it is known that SAEs trained on nearby layers in the residual stream learn very similar features, effectively wasting training compute, while SAEs trained on nearby FFNs learn disjoint sets of features \cite{balagansky2024mechanistic}. For this reason, we suggest that practitioners who are planning to train more than one sparse coder on a model should consider training transcoders on FFNs in lieu of SAEs on the residual stream.

\section{Future work}

\label{sec:future}

As we have shown, transcoders preserve more of a model's behavior and produce more interpretable latents. We believe skip connections let the transcoder avoid the redundant work of translating the linear map, letting it focus on learning important features. Future work may illuminate the role of the skip connection by comparing it to a learned or analytically derived affine approximation of the MLP component.

\citet{transcoders_paper} highlights the usefulness of transcoders for circuit detection.While we have not run experiments on circuit analysis like in that paper, we expect that skip transcoders to be better for reconstructing circuits thanks to their lower reconstruction error. It is unlikely that the skip connection impedes gradient-based circuit discovery: work like \cite{marks2024sparse} shows ways of incorporating linear skip connections into circuit discovery faithfully.

Transcoder and skip transcoder features may be used for steering, but we could not translate the unlearning and concept erasure benchmarks from \citet{karvonen2024saebench}, which require the latent to contain all information in any given residual stream position.

Transcoders also do not help improve the feature learning in SAEs the way new architectures like \citet{gao2024scaling} do. They merely change the objective of the SAE, which is something that cannot lessen inefficiencies in training. We see the effects of this in the evaluation results on feature absorption: transcoders and skip transcoders can exhibit it just as much as SAEs. Further, feature density plots do not exhibit significant differences \cref{sec:app-density}, showing that SAEs and SSTs are similar on a mechanistic level. Work like Matryoshka SAEs \cite{bussman2024matryoshka,nabeshima2024matryoshka} may help tackle these issues for both SAEs and SSTs.

\section{Contributions}

Nora Belrose had the idea that transcoders might be a superior architecture to SAEs, and came up with the idea of skip transcoders. Gonçalo Paulo trained most of the SAEs and transcoders, and performed the experiments and data analysis. Gonçalo and Stepan wrote the first draft. Nora Belrose, Stepan, and Gonçalo wrote the final version. Nora Belrose provided guidance and suggested experiments. Gonçalo, Nora, and Stepan are funded by a \href{https://www.openphilanthropy.org/grants/eleuther-ai-interpretability-research/}{grant} from Open Philanthropy. We thank Coreweave for computing resources.

\section{Code availability.}

Code for training transcoders and skip transcoders is available in the \href{https://github.com/EleutherAI/sparsify}{sparsify GitHub repo}. The skip transcoder checkpoints for Llama 3.2 1B are available on the HuggingFace Hub \href{https://huggingface.co/EleutherAI/skip-transcoder-Llama-3.2-1B-131k}{here}, and others will be uploaded to the Hub soon. 

\section*{Impact Statement}

This paper presents work whose goal is to advance the field of Mechanistic Interpretability. There are many potential societal consequences 
of our work, none which we feel must be specifically highlighted here.

\bibliography{bibliography}
\bibliographystyle{icml2025}

%%%%%%%%%%%%%%%%%%%%%%%%%%%%%%%%%%%%%%%%%%%%%%%%%%%%%%%%%%%%%%%%%%%%%%%%%%%%%%%
%%%%%%%%%%%%%%%%%%%%%%%%%%%%%%%%%%%%%%%%%%%%%%%%%%%%%%%%%%%%%%%%%%%%%%%%%%%%%%%
% APPENDIX
%%%%%%%%%%%%%%%%%%%%%%%%%%%%%%%%%%%%%%%%%%%%%%%%%%%%%%%%%%%%%%%%%%%%%%%%%%%%%%%
%%%%%%%%%%%%%%%%%%%%%%%%%%%%%%%%%%%%%%%%%%%%%%%%%%%%%%%%%%%%%%%%%%%%%%%%%%%%%%%
\newpage
\appendix
\onecolumn
\section{SAEBench results}
\label{sec:app-saebench}

This section contains results on SAEBench \cite{karvonen2024saebench}. We run three of the evaluations: \textit{core} (reconstruction quality), sparse probing and absorption. We describe all three in the main body and point out that it is not expected for transcoders to outperform SAEs on \textit{absorption}. 

\begin{table}[h]
\centering
\begin{tabular}{|c|cc|cc|cc|cc|}
\hline
 & \multicolumn{2}{c|}{\textbf{Variance Explained (\%)}} & \multicolumn{2}{c|}{\textbf{$\Delta$ NLL ($\downarrow$, \%)}} & \multicolumn{2}{c|}{\textbf{Sparse probing ($\uparrow$)}} & \multicolumn{2}{c|}{\textbf{Absorption score ($\downarrow$)}} \\
\textbf{Layer} & SAE & SST & SAE & SST & SAE & SST & SAE & SST \\
\hline
L10 & 16.5 & \textbf{67.1} & 1.1 & \textbf{0.5} & \textbf{71.5} & 70.6 & 36.3 $\pm$ 20.2 & \textbf{28.6 $\pm$ 13.3} \\
L14 & 17.0 & \textbf{72.4} & 1.1 & \textbf{0.5} & \textbf{80.0} & 76.0 & 33.1 $\pm$ 19.4 & \textbf{25.0 $\pm$ 18.4} \\
L18 & 20.9 & \textbf{81.7} & 2.1 & \textbf{0.5} & \textbf{78.9} & 75.2 & \textbf{26.3 $\pm$ 23.7} & 53.3 $\pm$ 19.6 \\
L22 & 21.6 & \textbf{73.2} & 1.6 & \textbf{0.5} & 76.1 &   & 24.4 $\pm$ 22.0 & \textbf{18.8 $\pm$ 29.5} \\
\hline
\end{tabular}
\caption{Results for gemma-2-2B. $\Delta$ NLL represents the increase in cross-entropy loss.}
\end{table}

\begin{table}[h]
\centering
\begin{tabular}{|c|ccc|ccc|ccc|ccc|}
\hline
 & \multicolumn{3}{c|}{\textbf{Variance Explained (\%)}} & \multicolumn{3}{c|}{\textbf{$\Delta$ NLL ($\downarrow$, \%)}} & \multicolumn{3}{c|}{\textbf{Sparse probing ($\uparrow$)}} & \multicolumn{3}{c|}{\textbf{Absorption score ($\downarrow$)}} \\
\textbf{Layer} & SAE & ST & SST & SAE & ST & SST & SAE & ST & SST & SAE & ST & SST \\
\hline
L0 & 99.4 & 99.4 & \textbf{99.9} & 0.2 & 0.2 & \textbf{0.0} & \textbf{71.9} & 66.1 & 69.3 & 95.5 $\pm$ 5.9 & \textbf{54.1 $\pm$ 10.5} & 60.3 $\pm$ 16.3 \\
L2 & 82.3 & 80.7 & \textbf{85.3} & 1.1 & 1.4 & \textbf{1.0} & 59.8 & \textbf{67.0} & 66.0 & \textbf{14.5 $\pm$ 14.3} & 47.4 $\pm$ 19.2 & 33.5 $\pm$ 14.8 \\
L4 & 75.9 & 74.3 & \textbf{87.4} & 1.1 & 1.2 & \textbf{0.7} & 67.7 & 67.3 & \textbf{69.4} & \textbf{19.3 $\pm$ 18.7} & 62.2 $\pm$ 26.2 & 40.5 $\pm$ 17.5 \\
L6 & 81.0 & 77.8 & \textbf{86.5} & 1.1 & 1.2 & \textbf{0.7} & 65.9 & 69.5 & \textbf{69.9} & \textbf{8.2 $\pm$ 15.8} & 15.6 $\pm$ 18.9 & 31.3 $\pm$ 25.7 \\
L8 & 87.8 & 85.2 & \textbf{90.3} & 1.1 & 1.3 & \textbf{0.9} & 68.3 & 67.4 & \textbf{71.6} & \textbf{18.1 $\pm$ 16.8} & 82.9 $\pm$ 23.6 & 45.1 $\pm$ 21.7 \\
L10 & 86.5 & 84.4 & \textbf{88.8} & 1.4 & 1.7 & \textbf{1.3} & 69.0 & 71.0 & \textbf{73.7} & 36.4 $\pm$ 28.9 & \textbf{30.5 $\pm$ 29.9} & 33.9 $\pm$ 25.1 \\
\hline
\end{tabular}
\caption{Results for pythia-160m. $\Delta$ NLL represents the increase in cross-entropy loss.}
\end{table}

\begin{table}[h]
\centering
\begin{tabular}{|c|cc|cc|cc|cc|}
\hline
 & \multicolumn{2}{c|}{\textbf{Variance Explained (\%)}} & \multicolumn{2}{c|}{\textbf{$\Delta$ NLL ($\downarrow$, \%)}} & \multicolumn{2}{c|}{\textbf{Sparse probing ($\uparrow$)}} & \multicolumn{2}{c|}{\textbf{Absorption score ($\downarrow$)}} \\
\textbf{Layer} & SAE & SST & SAE & SST & SAE & SST & SAE & SST \\
\hline
L0 & 93.0 & \textbf{93.8} & 1.5 & \textbf{1.0} & \textbf{69.9} & 69.5 & 80.7 $\pm$ 12.7 & \textbf{4.7 $\pm$ 2.8} \\
L2 & 73.8 & \textbf{80.5} & \textbf{1.5} & \textbf{1.5} & 69.5 & \textbf{72.4} & \textbf{30.2 $\pm$ 22.1} & 42.3 $\pm$ 5.1 \\
L4 & 63.3 & \textbf{81.2} & 1.5 & \textbf{1.0} & 70.9 & \textbf{74.5} & \textbf{46.7 $\pm$ 19.4} & 60.7 $\pm$ 27.5 \\
L6 & 57.8 & \textbf{78.9} & 1.5 & \textbf{1.0} & 68.7 & \textbf{69.9} & 82.6 $\pm$ 8.7 & \textbf{36.0 $\pm$ 14.6} \\
L8 & 63.3 & \textbf{82.8} & 1.5 & \textbf{1.0} & \textbf{72.2} & 69.9 & \textbf{66.1 $\pm$ 12.7} & 75.1 $\pm$ 10.3 \\
L10 & 69.9 & \textbf{84.4} & 2.0 & \textbf{1.5} & \textbf{74.9} & 73.6 & \textbf{44.7 $\pm$ 21.0} & 52.3 $\pm$ 16.2 \\
L12 & 71.1 & \textbf{77.0} & \textbf{2.0} & \textbf{2.0} &   & 74.6 & \textbf{9.4 $\pm$ 11.0} & 45.4 $\pm$ 18.9 \\
L14 & 71.1 & \textbf{75.8} & \textbf{2.0} & \textbf{2.0} & 71.1 & \textbf{75.3} & \textbf{0.1 $\pm$ 0.4} & 52.0 $\pm$ 19.9 \\
\hline
\end{tabular}
\caption{Results for llama-1B.$\Delta$ NLL represents the increase in cross-entropy loss.}
\end{table}
\newpage
\section{Feature density comparison}
\label{sec:app-density}

\begin{figure}[h]
    \centering    \includegraphics[width=1\linewidth]{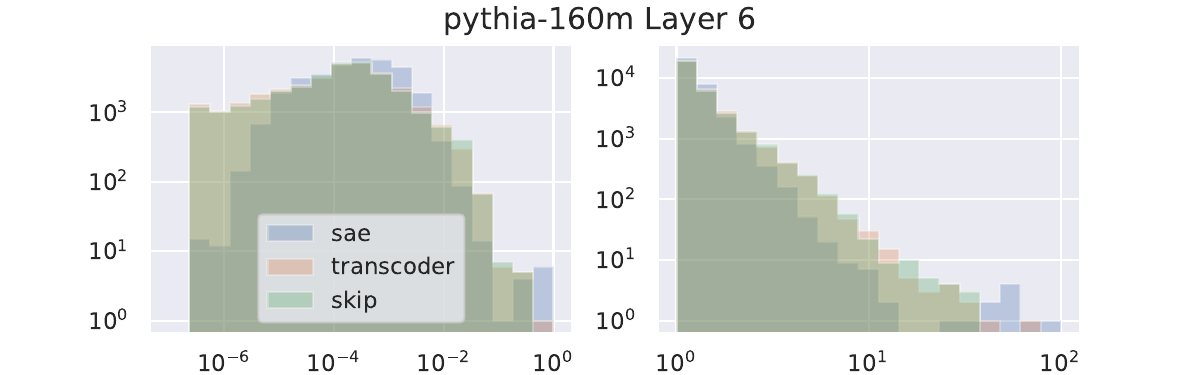}
    \centering    \includegraphics[width=1\linewidth]{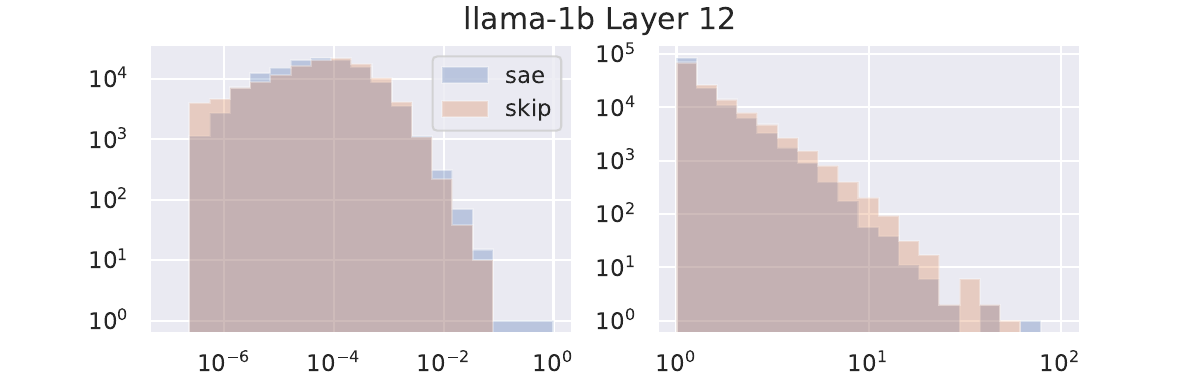}
    \centering    \includegraphics[width=1\linewidth]{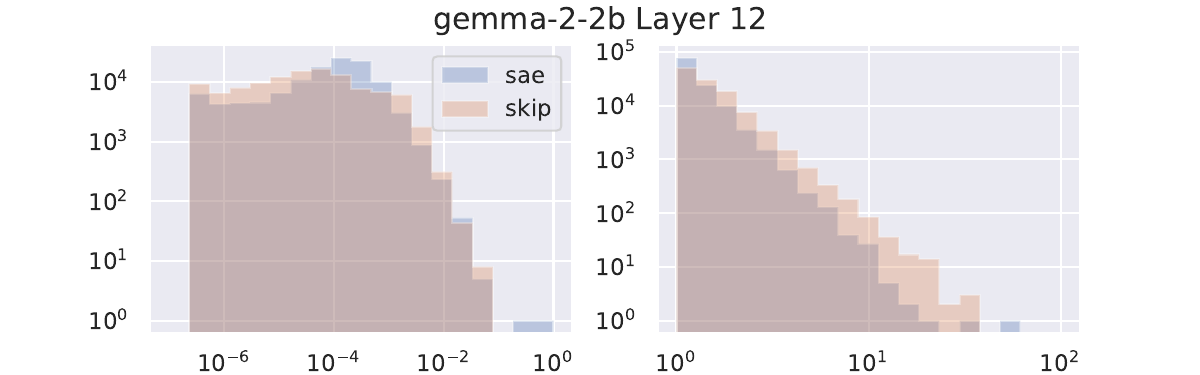}
    \vskip -0.3cm
    \caption{Comparison of feature density \cite{bricken2023towards} and the consistent activation heuristic (sum of activations over all tokens divided by the number of tokens). These plots show that STs and SSTs are similar in terms of feature density and have less high-density features and more low-density features. This is not a problem because there exist methods for getting rid of low-density features \citep{bricken2023towards,jermyn24ghostgrads,gao2024scaling}, but not for regularizing high-density features.}
    \label{fig:density}
\end{figure}

\end{document}